\documentclass[runningheads]{llncs}
\usepackage{xcolor, soul}
\usepackage{graphicx}
\usepackage{mwe}
\usepackage{cite}
\usepackage{amsfonts}
\usepackage[colorinlistoftodos]{todonotes}
\usepackage{url}
\usepackage{booktabs}
\newcommand{\ra}[1]{\renewcommand{\arraystretch}{#1}}

\begin{document}
\title{Synthetic patches, real images: screening for centrosome aberrations in EM images of human cancer cells}
\titlerunning{Synthetic patches, real images}
\author{Artem Lukoyanov\inst{1,2} \and
Isabella Haberbosch\inst{1,3} \and 
Constantin Pape\inst{1} \and
Alwin Kr{\"a}mer\inst{3} \and
Yannick Schwab\inst{1} \and
Anna Kreshuk\inst{1}}

\authorrunning{A. Lukoyanov et al.}
\institute{European Molecular Biology Laboratory (EMBL) \and Smart Engines, Moscow \and Clinical Cooperation Unit Molecular Hematology/Oncology, German Cancer Research Center (DKFZ)}
\maketitle              %
\begin{abstract}
Recent advances in high-throughput electron microscopy imaging enable detailed study of centrosome aberrations in cancer cells. While the image acquisition in such pipelines is automated, manual detection of centrioles is still necessary to select cells for re-imaging at higher magnification. In this contribution we propose an algorithm which performs this step automatically and with high accuracy. From the image labels produced by human experts and a 3D model of a centriole we construct an additional training set with patch-level labels. A two-level DenseNet is trained on the hybrid training data with synthetic patches and real images, achieving much better results on real patient data than training only at the image-level. The code can be found at \url{https://github.com/kreshuklab/centriole_detection}.

\end{abstract}
\section{Introduction}

Centrosomes consist of a pair of centrioles embedded in pericentriolar material and act as the major microtubule organizing centers of eukaryotic cells. They are pivotal for several fundamental cellular processes, including formation of bipolar mitotic spindles, a process essential for accurate chromosome segregation. Centrosome aberrations are a hallmark of cancer and have been described in virtually all malignancies examined. Surprisingly, with very few exceptions, no electron microscopy (EM) data on aberrant centrosomes in primary tumor tissues are available. As a consequence, next to nothing is known with regard to ultrastructure, biogenesis and functional consequences of specific types of centrosome aberrations. The main roadblock of such studies has long been on the image acquisition side, since high-throughput EM imaging is technically very difficult to orchestrate. The situation has recently been alleviated by the introduction of software tools for targeted electron microscopy \cite{schorb2018software}, which allow for automated imaging of a large number (1000-2000) of cells on thin resin sections. These cells then need to be screened for centrosome/centriole abnormalities. The screening - a time consuming, tedious task with a hit rate of 2 to 5 \% - is currently performed manually. This contribution proposes an algorithm for automatic centriole detection and thus closes the last methodological gap on the way to fully automated screening for centrosome aberrations within a cancer cell population.

\begin{figure}
    \centering
    \minipage{0.32\textwidth}
        \includegraphics[width=\linewidth]{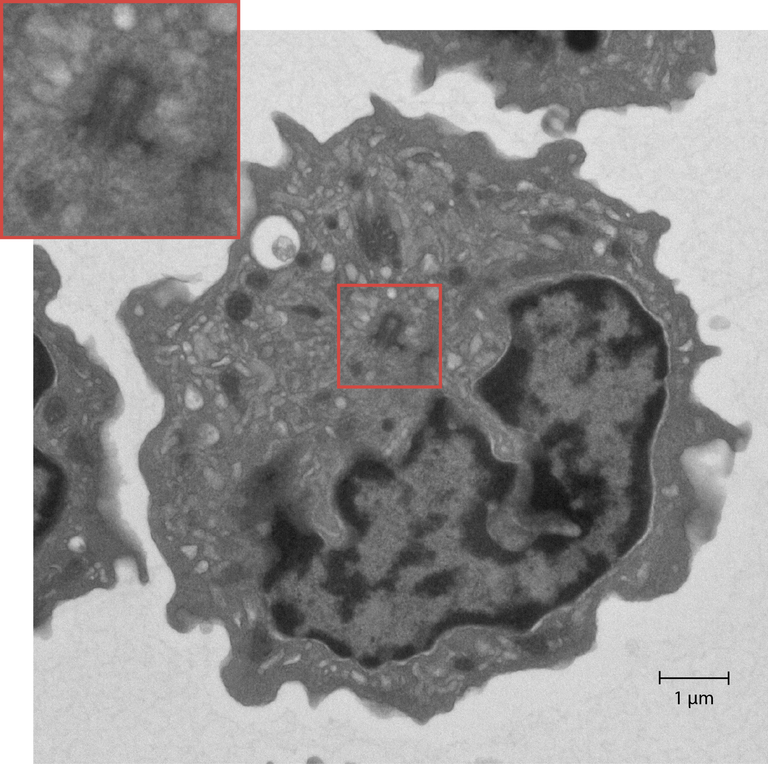}
    \endminipage\hfill
    \minipage{0.32\textwidth}
        \includegraphics[width=\linewidth]{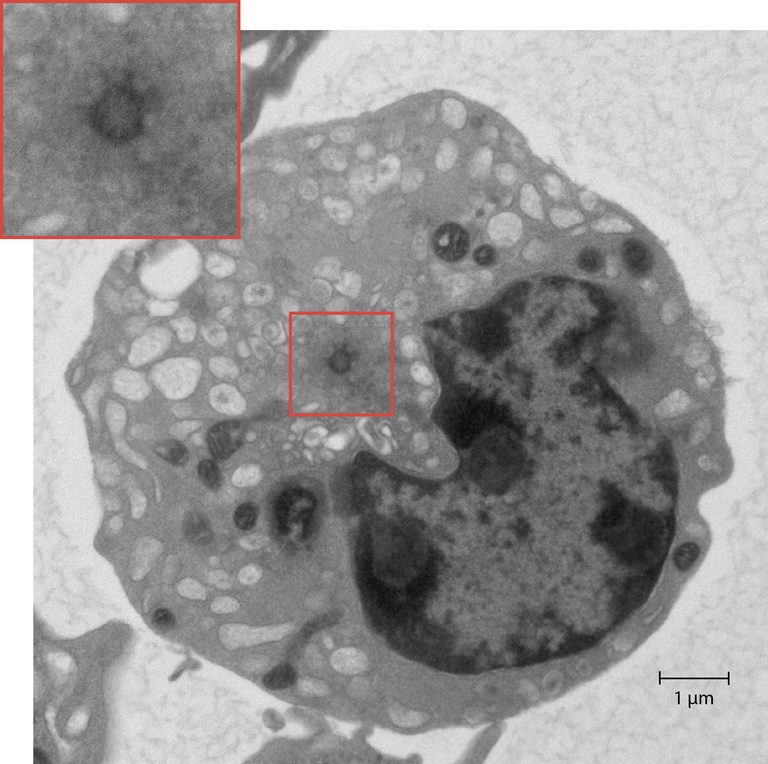}
    \endminipage\hfill
    \minipage{0.32\textwidth}
        \includegraphics[width=\linewidth]{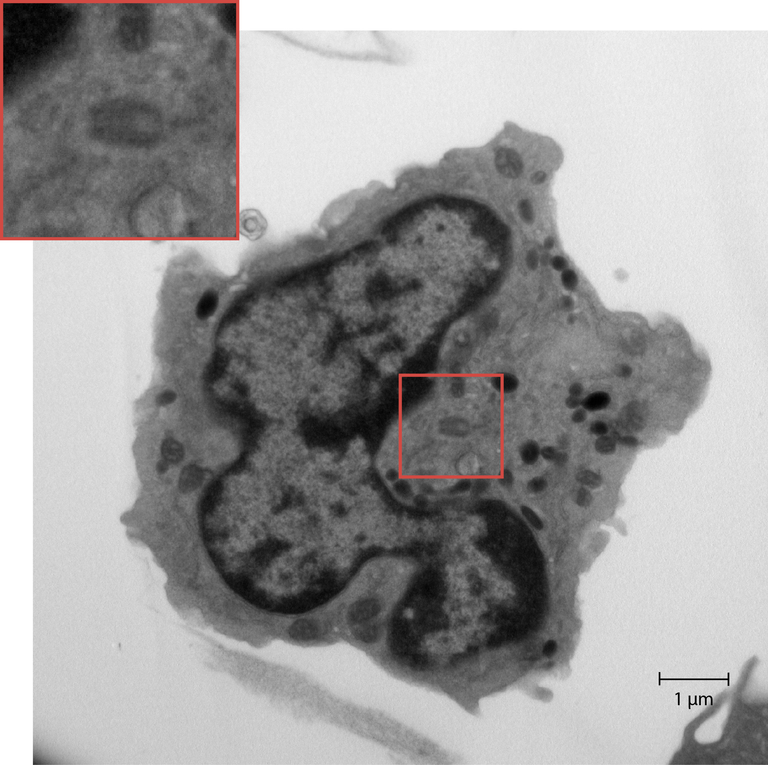}
    \endminipage
    \vskip\baselineskip
    \minipage{0.32\textwidth}
        \includegraphics[width=\linewidth]{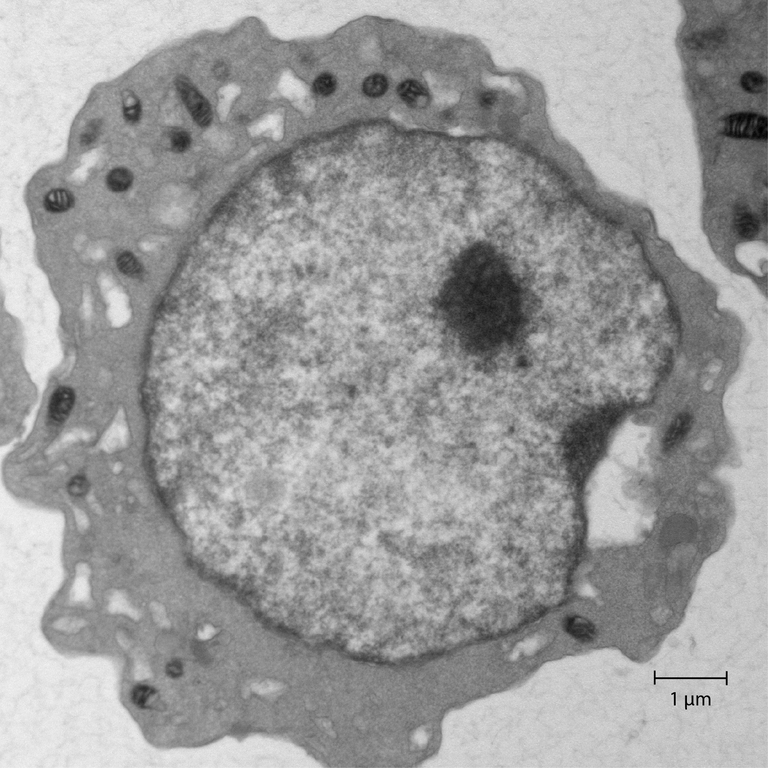}
    \endminipage\hfill
    \minipage{0.32\textwidth}
        \includegraphics[width=\linewidth]{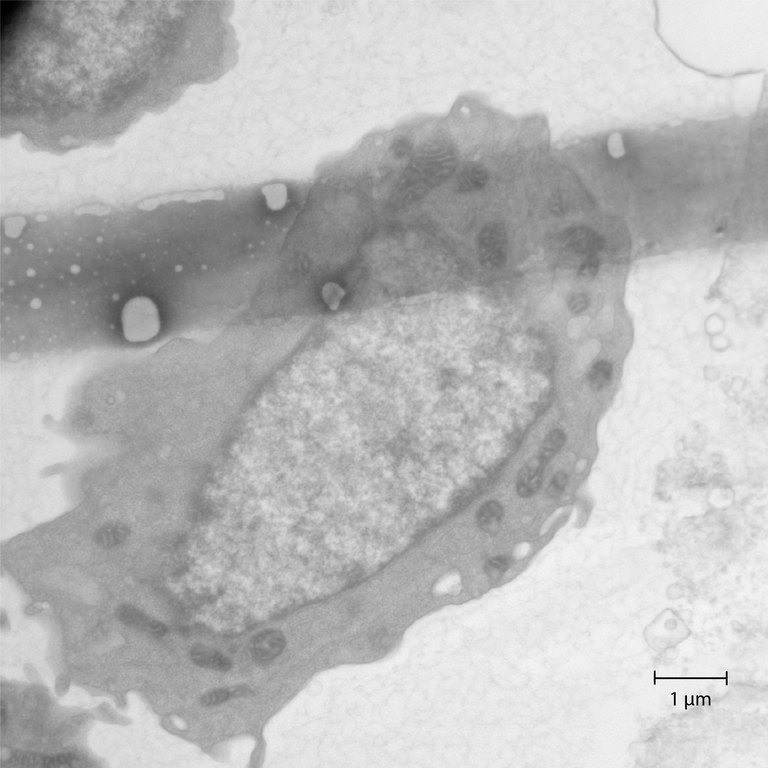}
    \endminipage\hfill
    \minipage{0.32\textwidth}
        \includegraphics[width=\linewidth]{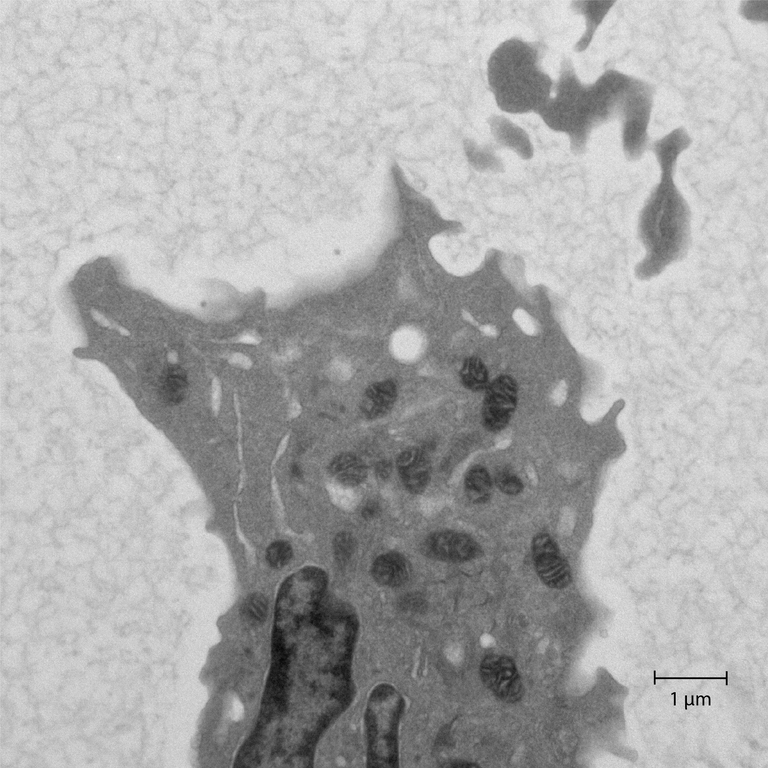}
    \endminipage
    \caption{\small Examples of positive (top) and negative (bottom) images from the screening. The centrioles of the positive images are shown in more detail in the top left corners.}
    \label{fig:centrioles_samples}
\end{figure}

During screening, experts mark which images contain a centriole to trigger re-imaging at higher magnification or with TEM tomography for detailed study of the ultrastructure. The task we intend to solve is thus one of image classification, as no labels are placed on the centrioles themselves and their location in the positive images is not known. Fig.~\ref{fig:centrioles_samples} shows several example images from the screening, illustrating the challenges of the classification task. Centrioles are fairly small and, to an untrained eye, look very similar to other ultrastructure elements. Besides, since cell sectioning is done without seeing the centrioles, they can be positioned at any orientation to the cutting plane. Consequently, their appearance - roughly cylindrical in 3D - can range from a circle to two parallel lines (Fig.~\ref{fig:centrioles_samples}, top row). Finally, the screening images are not artifact-free (Fig.~\ref{fig:centrioles_samples}, bottom center) and can differ significantly in their intensity distribution (Fig.~\ref{fig:centrioles_samples}, bottom left and right). 

The task of automated centriole detection has, to the best of our knowledge, not been addressed before in EM images. The state-of-the-art algorithms for detection of other objects in EM images rely on dense object-level labels \cite{heinrich2018synaptic, beier2017multicut, xiao2018automatic} as input for training a convolutional neural network (CNN). We intend to use a CNN as well, but we are limited to much weaker image-level labels. Numerous image classification CNNs developed for natural image processing could provide a good fit, but did not work in practice. Their lack of performance can perhaps be explained by the relatively very small size of the object in question. Specialized algorithms for the detection of small objects have been proposed in the remote sensing domain \cite{pang2019cnn, yang2018automatic}, but they again rely on object-level labels. 

Besides direct image classification, the screening problem can also be addressed in the framework of multiple instance learning \cite{kandemir2015computeraided}. This approach is often used in computer-aided diagnostics, with image patches serving as instances and complete images as bags. The main drawback for our application is the intentional removal of spatial context between the patches. %

Artificial image generation is a popular means to augment insufficient training data, used both in medical \cite{mahmood2018unsupervised} and natural image domains \cite{tremblay2018training}. For object detection in particular, several suggestions have been made on how an object can be consistently pasted into an empty background scene \cite{gupta2016synthetic, dwibedi2017cut}. Inspired by this work, we propose to limit the synthesis to image patches and create synthetic groundtruth by pasting slices of a 3D centriole model into patches of a few negatively labeled training images. The combined hybrid groundtruth of synthetic patches and real images is used to train a special two-level neural network. The training begins from the first patch-level sub-network. After this network is trained (on synthetic patches), we freeze its weights and attach the image-level sub-network to be trained on real images.

The hybrid groundtruth of synthetic patches and real images has several advantages over existing approaches: i) we avoid the difficult task of generating whole cell images with correct placement of all the organelles; ii) we can train a powerful network on the patch level and capture all the fine ultrastructure in the patch necessary to recognize centrioles  from other similar objects; iii) we can preserve spatial context between the patches for the image-level prediction.

Our complete model and training procedure are described in detail in the next section. In section~\ref{ref:sec_experiments}, we apply the network to real patient data and show that it achieves sufficiently high accuracy for fully automated screening. Finally, in section~\ref{sec_discussion} we discuss other possible applications of our hybrid training setup.

\section{Methods}
\subsection{Generation of synthetic patches}
\begin{figure}[!htb]
\minipage{0.12\textwidth}
  \includegraphics[width=\linewidth]{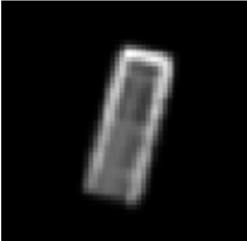}
\endminipage\hfill
\minipage{0.12\textwidth}
  \includegraphics[width=\linewidth]{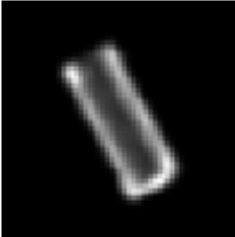}
\endminipage\hfill
\minipage{0.12\textwidth}%
  \includegraphics[width=\linewidth]{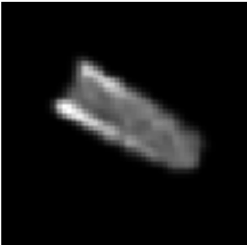}
\endminipage\hfill
\minipage{0.12\textwidth}%
  \includegraphics[width=\linewidth]{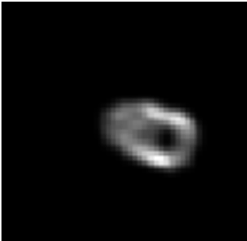}
\endminipage\hfill
\minipage{0.12\textwidth}%
  \includegraphics[width=\linewidth]{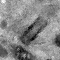}
\endminipage\hfill
\minipage{0.12\textwidth}%
  \includegraphics[width=\linewidth]{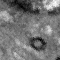}
\endminipage\hfill
\minipage{0.12\textwidth}%
  \includegraphics[width=\linewidth]{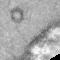}
\endminipage\hfill
\minipage{0.12\textwidth}%
  \includegraphics[width=\linewidth]{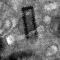}
\endminipage
\caption{Left: artificially generated slices of the cylinder model of a centriole. Right: model slices painted into background cell patches.}
\label{fig:slices_image}
\end{figure}

We start from a simplified 3D model of a centriole which at this resolution can be described as two orthogonal hollow cylinders roughly 250nm in diameter and 500nm in length. Transmission Electron Microscopy works by detecting electrons which pass through a sample - in this case, a 200nm section of a cell. To simulate this process, we perform three random rotations of the model around three axes, pick a random length value and take a 10-pixel slice out of the model at this position. The slice is then summed into a 2D image along the depth axis and Gaussian blur is applied. Our 3D model and sample images are shown in Fig.~\ref{fig:slices_image}.

The generated model slices need to be combined with realistic cell background patches. Instead of generating them ad-hoc, we select patches from the negative training samples since those were labeled not to contain a centriole. The screening images contain multiple cells and the negative label of the whole image only refers to the central one. We select the central cell by a combination of smoothing, binarization, morphological erosion, connected components filtering and dilation. Since large intensity shifts are present between the screening images, we choose an adaptive binarization threshold of $0.9 \times \mathbb{E}(I)$, where $\mathbb{E}(I)$ stands for average intensity of the image.

Once the cell image without centrioles is selected, we proceed to choose a patch inside the cell. We prefer relatively empty patches without other organelles in the middle, as we do not want to paste the centriole on top of a different object. We generate a set of $60\times60$ boxes covering the cell and randomly select from the patches of this set, weighing the probability for the patch to be selected by the standard deviation of the intensity inside it: $P_{selected} = 1/({N_{patches} * \sigma^4)}$.

The final step consists of painting the model slice into the background patch. Since image formation in TEM depends on the scattering of electrons while they pass through the specimen, we propose to subtract the model slice image from the background patch. Additionally, the generated patch needs to remain within the real image intensity range. Let $I_{bg}$ be the background image patch, $S$ the model slice, $Q_{5}$ the 5\% quantile. We combine the background and the model slice into the final image $I$ by the following formula:
$$ I = \textrm{max}(Q_5(I_{bg}), ~ I_{bg} - \alpha * \sigma * S * \epsilon)$$
Here, $\alpha$ is a scaling factor, $\sigma$ the standard deviation of the background patch and $\epsilon$ a random number between $0.9$ and $1.1$. A few examples of generated patches are shown in Fig.~\ref{fig:slices_image}.

\subsection{Neural network and hybrid training}
We use the synthetic patches described above to train a convolutional neural network for patch classification. Once the training on synthetic patches is finished, we incorporate the trained network into a bigger one which is then trained on real images with image-level labels.  

On the patch level, we use a DenseNet \cite{huang2017densely} with a growth rate of 32, three dense blocks with depths 6, 12 and 32, three fully connected layers with 1280, 80 and 16 input channels.   

\begin{figure}[!htb]
\minipage{\textwidth}
 \includegraphics[width=\linewidth]{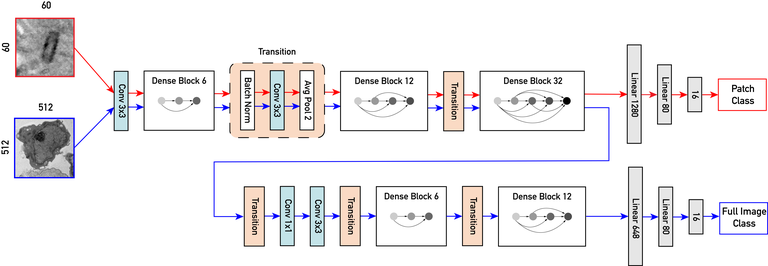}
\endminipage
\caption{The architecture of the proposed two-level network. The full network is based on DenseNet-B, i.e. a DenseNet with a bottleneck $1\times1$ convolution layer. Activation layers are omitted for clarity. Code will be made publicly available.}
\label{fig:architecture}
\end{figure}

The network architecture is illustrated in Fig.~\ref{fig:architecture}. We construct the image-level network by extending the patch-level one. In more detail, we keep the first three dense blocks, remove the fully connected layers, add two more convolutional layers and two more dense blocks of depth 6 and 12, and finally add three fully connected layers with 648, 80 and 16 channels. The weights of the first three dense blocks are frozen at the values that were obtained during patch-level training. The new layers of the network are trained on real images for the task of image classification. After the training converges, we un-freeze the weights of the first three blocks and continue the training end-to-end until convergence. 

Our motivation for the 3-step training procedure is as follows: the goal of the first training step is to tune the low-level filters in the first blocks to the problem of centriole detection. The second step learns to combine the patch-level filter responses with the global image context. Finally, the third step trains the complete network end-to-end and learns to recognize centrioles in real images and classify them even better. 

Besides the hybrid patch/image model above, we investigate two simpler alternatives. Both start from the patch-level network trained on synthetic data. We apply the network to the whole image in a sliding window manner with a stride of 10 pixels. For each pixel we obtain a score for the probability of the patch centered at this pixel to contain a centriole. In the first approach we then threshold the probabilities and assign the image a positive label if at least one area of the image remains above the threshold (there is at least one strong centriole detection). The second approach uses the probability map as input to another neural network.
Note that unlike the hybrid network neither of these methods allow for retraining the patch-level neural network.

\section{Experiments}
\label{ref:sec_experiments}
\subsection{Dataset description}
For imaging in TEM, patient cells are collected, fixed and pelleted. The pellets are then prepared for electron microscopy, embedded in resin and cut at random positions as 200 nm thick sections. Typically, one section contains about 1 to 2 thousands cell cross-sections, at variable positions. In the case of blast cells, their average diameter is 8-15 $\mu m$; even though each cell should contain at least one pair of centrioles the typical centriole dimensions being 200 nm by 500 nm, the probability to hit one is from 2 to 5\%. Each cell cross-section present on the EM grid is imaged automatically at a magnification allowing for unambiguous identification of centriole profiles. The resulting stack of about 1 to 2 thousands images is then screened manually by an expert for selecting the cells that show a centriole on that specific section. 

In total the dataset contains 742 positive images and approximately 16 000 negative images, from 6 to 147  positive images for each of 14 patients. For each patient, we randomly sample the negative images to obtain a balanced training and test set. The performance of the screening algorithms is measured as accuracy of image label assignment.

\subsection{State-of-the-art image classification models}
First we apply the standard image classification networks, namely VGG\cite{simonyan2015very}, ResNet\cite{he2016deep}, DenseNet and DenseNet with additional skip connections between dense blocks. The best result on a balanced test set was at 58\% accuracy. All models overfit very quickly even when regularization is used. Augmenting the dataset with synthetic images generated by the same procedure as we use for patches did not improve the performance.

The next experiment was based on multiple instance learning, where we implemented the attention-based model of \cite{ilse2018attentionbased}. Since at most one centriole is visible, the positive bags only had a single positive instance and the network never achieved accuracy above 54\%. 

\subsection{Hybrid training}
\begin{figure}
    \centering
        \includegraphics[width=0.24\linewidth]{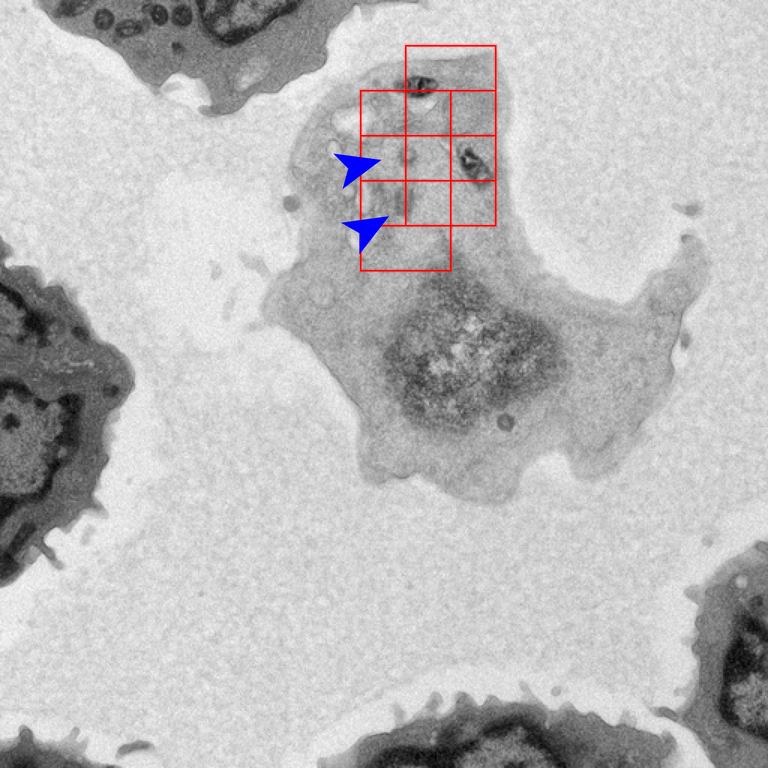}
        \includegraphics[width=0.24\linewidth]{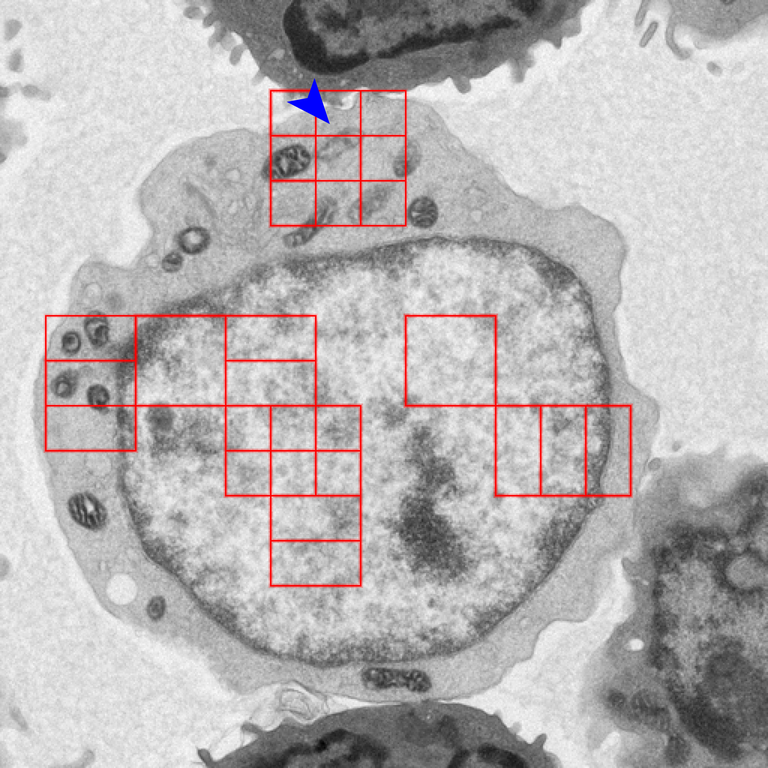}
        \includegraphics[width=0.24\linewidth]{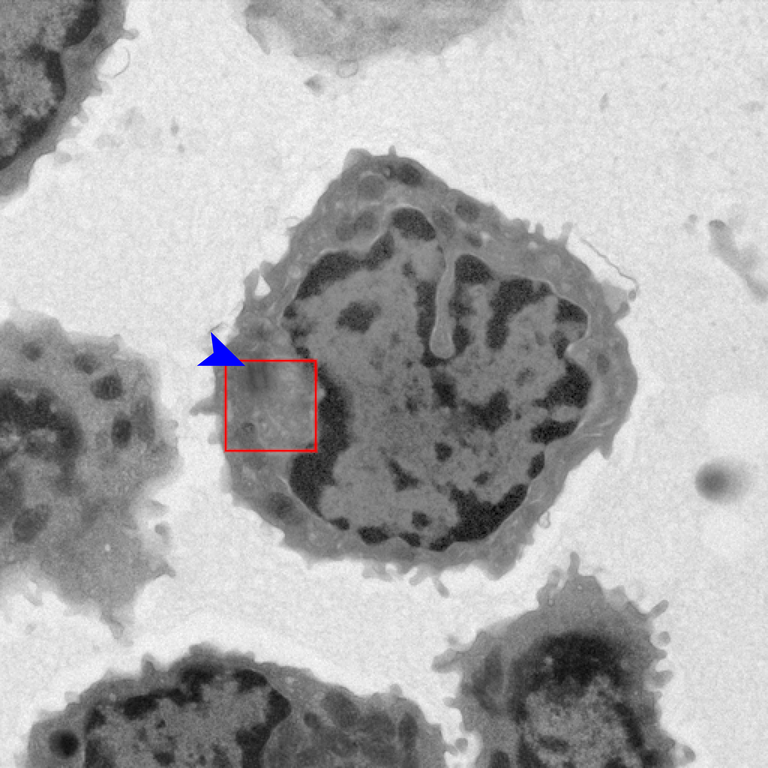}
        \includegraphics[width=0.24\linewidth]{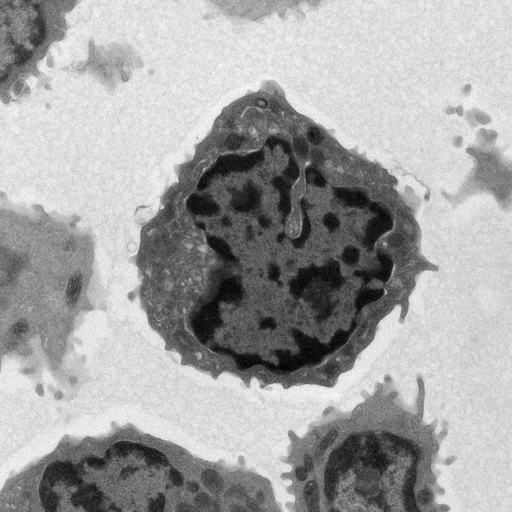}
\caption{Real images with patch network predictions (red boxes) and real centrioles positions (blue arrowheads). The first 3 images contain centrioles which are detected correctly. In the 2nd image false positive detections are present along with the correct centriole patch. The 4th image does not contain a centriole and is correctly predicted as negative. }
    \label{fig:patch_predictions}
\end{figure}

The patch-level network operated on 60$\times$60 patches. In total, approximately 20 000 (18 000 for train and 2000 for test) patches were generated for its training in each experiment. The network achieved the accuracy of 98.7\% on synthetic patches. A few examples of its predictions on real images are shown in Fig.~\ref{fig:patch_predictions}. The fourth image is correctly predicted as negative and the rest as positive. However, spurious predictions as shown in the second image are often present at objects similar to centrioles, and also inside the nucleus which resembles the empty background we used in the positive samples.

We compare several approaches to aggregating patch-level information for image-level prediction. In the first one, the patch level predictions are thresholded at 50\% and the whole image is assigned to positive class if at least one patch is positive. In the second, we train a DenseNet on the predictions of the patch network. 
For the proposed hybrid training, we evaluate a 2-step and a 3-step training procedure. In the 2-step training the weights of the first three dense blocks remain frozen after the training of the patch-level sub-network. We perform 4-fold cross validation on the patient level, using 75\% of the images from 9 patients for training and the other for validation. Five unseen patients are used for testing. The results on different folds are very consistent. As the final experiment, we unfreeze the weights of the first three dense blocks for the network trained on one of the folds and continue training end-to-end until convergence. The results of our experiments are summarized in Table~\ref{table:results}.

The hyperparameters were set as follows: Adam optimizer with $betas = (0.9, 0.999)$, $eps = 10^{-8}$, $learning~rate = 10^{-3}$ and scheduler which reduces learning rate by 5\% each 10 epochs if the validation does not improve. We perform the standard data augmentation: random flips, random translation at 10\%, random re-scale at 10\%, random rotation around arbitrary axes.

\begin{table}[t]
\centering
\ra{1.3}
\caption{Accuracy of the algorithms on a balanced test set from unseen patients}
\begin{tabular}{@{}p{2.8cm}p{2.8cm}p{3cm}p{3cm}@{}}

\hline
Threshold patch & Train from & Hybrid network & Hybrid network \\
predictions & patch predictions & 2-step training & 3-step training \\
\hline
68 & 64 & 80.36 $\pm$ 1.18 & 87.23 \\
\hline

\end{tabular}

\label{table:results}
\end{table}

\section{Discussion and future work}
\label{sec_discussion}

We have proposed an algorithm for automated centriole screening in EM images which can be trained, by using a 3D model of a centriole to generate synthetic patches, purely from image-level manual labels. On a test dataset of real patient data the algorithm demonstrates high accuracy and stability across folds in cross-validation. High-throughput screening is fairly tolerant to detection errors: as long as a sufficient number of centrioles are detected to draw statistical conclusions, a small number of false positives or false negatives are considered a reasonable price to pay for reduction in manual labor. The accuracy achieved by our algorithm was judged to be sufficient for fully automated screening. 

The main drawback of our approach comes from its main strength - it relies on the 3D model of a centriole to train the patch-level classifier. We show in Table \ref{table:results} that the image-level classifier can learn to correct patch-level errors. Still, centrosome aberrations which lead to a significant change in centriole appearance will likely be confusing for the algorithm. These, however, would be challenging for a human expert as well. 

Outside the original application of screening for centriole presence, the hybrid training approach would apply equally well to other organelles of uniform shape, such as various coated vesicles. At even higher resolution it can be extended to large single molecules, while at much lower resolution in the natural image domain it can be applied to problems of defect detection and quality control.

\bibliographystyle{splncs04}
\bibliography{Centriole_paper}

\end{document}